# PRINCIPAR—An Efficient, Broad-coverage, Principle-based Parser

Dekang Lin

Department of Computer Science, University of Manitoba

Winnipeg, Manitoba, Canada R3T 2N2, lindek@cs.umanitoba.ca

**Abstract**

We present an efficient, broad-coverage, principle-based parser for English. The parser has been implemented in C++ and runs on SUN Sparc-stations with X-windows. It contains a lexicon with over 90,000 entries, constructed automatically by applying a set of extraction and conversion rules to entries from machine readable dictionaries.

## 1. Introduction

Principle-based grammars, such as Government-Binding (GB) theory (Chomsky, 1981; Haegeman, 1991), offer many advantages over rule-based and unification-based grammars, such as the universality of principles and modularity of components in the grammar. Principles are constraints over X-bar structures. Most previous principle-based parsers, e.g., (Dorr, 1991; Fong, 1991; Johnson, 1991), essentially generate all possible X-bar structures of a sentence and then use the principles to filter out the illicit ones. The drawback of this approach is the inefficiency due to the large number of candidate structures to be filtered out. The problem persists even when various techniques such as optimal ordering of principles (Fong, 1991), and coroutining (Dorr, 1991; Johnson, 1991) are used. This problem may also account for the fact that these parsers are experimental and have limited coverage.

This paper describes an efficient, broad-coverage, principle-based parser, called PRINCIPAR. The main innovation in PRINCIPAR is that it applies principles to descriptions of X-bar structures rather than the structures themselves. X-bar structures of a sentence are only built when their descriptions have satisfied all the principles.

Figure 1 shows the architecture of PRINCIPAR. Sentence analysis is divided into three steps. The lexical analyser first converts the input sentence into a set of lexical items. Then, a message passing algorithm for GB-parsing is used to construct a shared parse forest. Finally, a parse tree retriever is used to enumerate the parse trees.

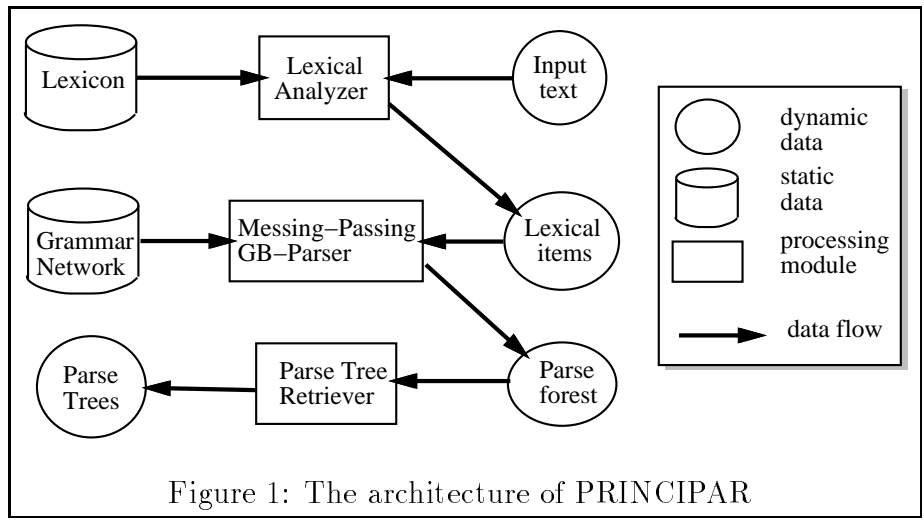

Figure 1: The architecture of PRINCIPAR

The key idea of the parsing algorithm was presented in (Lin, 1993). This paper presents some implementation details and experimental results.

## 2. Parsing by Message Passing

The parser in PRINCIPAR is based on a message-passing framework proposed by Lin (1993) and Lin and Goebel (1993), which uses a network to encode the grammar. The nodes in the grammar network represent grammatical categories (e.g., NP, Nbar, N) or subcategories, such as V:NP (transitive verbs that take NPs as complements). The links in the network represent relationships between the categories. GB-principles are implemented as **local constraints** attached to the nodes and **percolation constraints** attached to links in the network. Figure 2 depicts a portion of the grammar network for English.

There are two types of links in the network: **subsumption links** and **dominance links**.

- There is a subsumption link from $\alpha$ to $\beta$ if $\alpha$ subsumes $\beta$. For example, since V subsumes V:NP and V:CP, there is a subsumption link from V to each one of them.

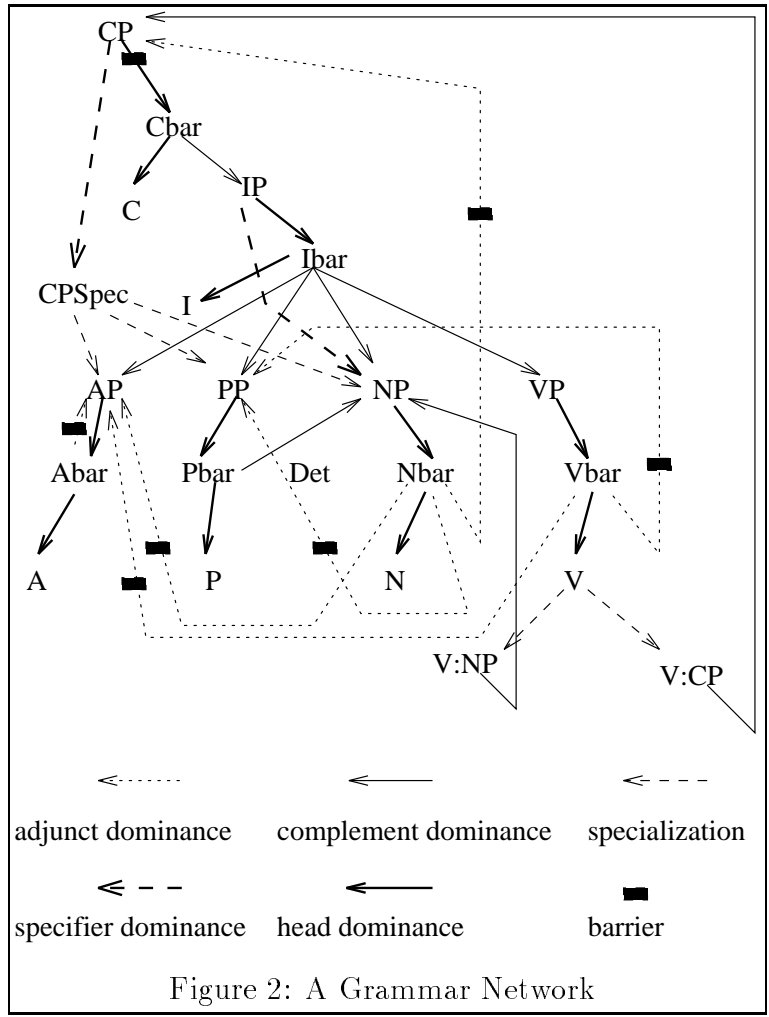

Figure 2: A Grammar Network

- There is a dominance link from node $\alpha$ to $\beta$ if $\beta$ can be immediately dominated by $\alpha$. For example, since an Nbar may immediately dominate a PP adjunct, there is a dominance link from Nbar to PP.

A dominance link from $\alpha$ to $\beta$ is associated with an integer id that determines the linear order between $\beta$ and other categories dominated by $\alpha$, and a binary attribute to specify whether $\beta$ is optional or obligatory.[1]

---

[1] In order to simplify the diagram, we did not label the links with their ids in Figure 2. Instead, the precedence between dominance links is indicated by their starting points, e.g, C precedes IP under Cbar since the link leading to C is to the left of the link leading to IP.

Input sentences are parsed by passing messages in the grammar network. The nodes in the network are computing agents that communicate with each other by sending messages in the reverse direction of the links in the network. Each node has a local memory that stores a set of items. An item is a triplet that represents a (possibly incomplete) X-bar structure $\alpha$:

$$<\texttt{str, att, src}>, \text{where}$$

`str` is an integer interval [i,j] denoting the i'th to j'th word in the input sentence; `att` is the attribute values of the root node of the X-bar structure; and `src` is a set of source messages from which this item is combined. The source messages represent immediate constituents of the root node. Each node in the grammar network has a completion predicate that determines whether an item at the node is "complete," in which case the item is sent as a message to other nodes in the reverse direction of the links.

When a node receives an item, it attempts to combine the item with items from other nodes to form new items. Two items

$$<[i_1,j_1], A_1, S_1> \text{ and } <[i_2,j_2], A_2, S_2>$$

can be combined if

1. their surface strings are adjacent to each other: $i_2 = j_1+1$.
2. their attribute values $A_1$ and $A_2$ are unifiable.
3. the source messages come via different links: $\text{links}(S_1) \cap \text{links}(S_2) = \emptyset$, where links(S) is a function that, given a set of messages, returns the set of links via which the messages arrived.

The result of the combination is a new item:

$$<[i_1,j_2], \text{unify}(A_1, A_2), S_1 \cup S_2>.$$

The new item represents a larger X-bar structure resulting from the combination of the two smaller ones. If the new item satisfies the local constraint of the node it is considered valid and saved into the local memory. Otherwise, it is discarded. A valid item satisfying the completion predicate of the node is sent further as messages to other nodes.

The input sentence is parsed in the following steps.

**Step 1: Lexical Look-up:** Retrieve the lexical entries for all the words in the sentence and create a lexical item for each word sense. A lexical item is a triple: $<[i,j], \text{av}_{\text{self}}, \text{av}_{\text{comp}}>$, where [i,j] is an interval denoting the position of the word in the sentence; $\text{av}_{\text{self}}$ is the attribute values of the word sense; and $\text{av}_{\text{comp}}$ is the attribute values of the complements of the word sense.

**Step 2: Message Passing:** For each lexical item $<[i,j], \text{av}_{\text{self}}, \text{av}_{\text{comp}}>$, create an initial message <[i,j], av$_{\text{self}}$, ∅> and send this message to the grammar network node that represents the category or subcategory of the word sense. When the node receives the initial message, it may forward the message to other nodes or it may combine the message with other messages and send the resulting combination to other nodes. This initiates a message passing process which stops when there are no more messages to be passed around. At that point, the initial message for the next lexical item is fed into the network.

**Step 3: Build a Shared Parse Forest** When all lexical items have been processed, a shared parse forest for the input sentence can be built by tracing the origins of the messages at the highest node (CP or IP), whose `str` component is the whole sentence. The parse forest consists of the links of the grammar network that are traversed during the tracing process. The structure of the parse forest is similar to (Billot and Lang, 1989) and (Tomita, 1986), but extended to include attribute values.

The parse trees of the input sentence can be retrieved from the parse forest one by one. The next section explains how the constraints attached to the nodes and links in the network ensure that the parse trees satisfy all the principles.

## 3. Implementation of Principles

GB principles are implemented as local and percolation constraints on the items. Local constraints are attached to nodes in the network. All items at a node must satisfy the node's local constraint. Percolation constraints are attached to the links in the network. A message can be sent across a link only if the item satisfies the percolation constraint of the link.

We will only use two examples to give the reader a general idea about how GB principles are interpreted as local and percolation constraints. Interested reader is referred to Lin (1993) for more details.

### 3.1. Bounding Theory

The Bounding Theory (Subjancency) states that a movement can cross at most one barrier without leaving an intermediate trace. An attribute named `whbarrier` is used to implement this principle. A message containing the attribute value `-whbarrier` is used to represent an X-bar structure containing a position out of which a wh-constituent has moved, but without yet crossing a barrier. The value `+whbarrier` means that the movement has already crossed one barrier. Certain dominance links in the network are designated as barrier links. Bounding condi-tion is implemented by the percolation constraints attached to the barrier links, which block any message with `+whbarrier` and change `-whbarrier` to `+whbarrier` before the message is allowed to pass through.

### 3.2. Case Theory

Case Theory requires that every lexical NP be assigned an abstract case. The implementation of case theory in PRINCIPAR is based on the following attribute values: `ca, govern, cm`.

| | |
|---|---|
| `+ca` | the head is a case assigner |
| `-ca` | the head is not a case assigner |
| `+govern` | the head is a governor |
| `-govern` | the head is not a governor |
| `-cm` | an NP m-commanded by the head needs case marking |

The case filter is implemented as follows:

1. Local constraints attached to the nodes assign `+ca` to items that represent X-bar structures whose heads are case assigners (P, active V, and tensed I).

   | Node | Local Constraint |
   |---|---|
   | P | assign `+ca` to every item |
   | V | assign `+ca` to items with `-passive` |
   | I | assign `+ca` to items with `tense` attribute |

2. Every item at NP node is assigned an attribute value `-cm`, which means that the NP represented by the item needs to be case-marked. The `-cm` attribute then propagates with the item as it is sent to other nodes. This item is said to be the origin of the `-cm` attribute.

3. Barrier links do not allow any item with `-cm` to pass through, because, once the item goes beyond the barrier, the origin of `-cm` will not be governed, let alone case-marked.

4. Since each node in X-bar structure has at most one governor, if the governor is not a case assigner, the node will not be case-marked. Therefore, a case-filter violation is detected if `+govern -cm -ca` co-occur in an item. On the other hand, if `+govern +ca -cm` co-occur in an item, then the head daughter of the item governs and case-marks the origin of `-cm`. The case-filter condition on the origin of `-cm` is met. The `-cm` attribute is cleared. The local constraints attached to all the nodes check for the co-occurrences of `ca`, `cm`, and `govern` to ensure case-filter is not violated by any item.

## 4. Lexicon

The lexicon in PRINCIPAR consists of two hash tables: a primary one in memory and a secondary one on disk. The secondary hash table contains over 90,000 entries, most of which are constructed automatically by applying a set of extraction and conversion rules to entries in Oxford Advanced Leaner's Dictionary and Collins English Dictionary.

When a word is looked up, the primary hashtable is searched first. If an entry for the word is found, the lexical search is done. Otherwise, the secondary hash table is searched. The entry retrieved from the secondary table is inserted into the primary one, so that when the word is encountered again only in-memory search will be necessary.

The primary hash table is loaded from a file at the system start-up. The file also serves as a buffer for changes to the secondary hash table. When a lexical entry is added or modified, it is saved in the file for the primary hash table. The entry in the secondary hash table remains unchanged. Since the primary hash table is always consulted first, its entries override the corresponding entries in the secondary table. The reason why the buffer is needed is that the secondary hash table is designed in such a way that update speed is sacrificed for the sake of efficient retrieval. Therefore, updates to the secondary hash table should be done in batch and relatively infrequently.

The two-tier organization of the lexicon is transparent to the parser. That is, as far as the parser is concerned, the lexicon is an object that, given a word or a phrase, returns its lexical entry or `nil` if the entry does not exist in the lexicon. Lexical retrieval is very efficient, with over 90,000 entries, the average time to retrieve an entry is 0.002 second.

### 4.1. Lexical Entries

Although the lexicon currently used in PRINCIPAR contains only syntactic information, it may also be used to hold other types of information. Each lexical entry consists of an entry word or phrase and a list of functions with arguments:

```
(<entry-word-or-phrase>
  (<func-name> <arg> ... <arg>)
  (<func-name> <arg> ... <arg>)
```

```
        ... ...
        (<func-name> <arg> ...  <arg>))
```

For example,
```
(acknowledge
 (subcat ((cat v)) (((cat i) -bare_inf)))
 (subcat ((cat v)) (((cat n) (case acc))))
 (subcat ((cat v)) (((cat c))))
```
The function `subcat` returns a subcategorization frame of the word. The first argument of the function is the attribute values of the word itself. The second argument of the function is a list of attribute value vector for the complements of the word. For example, the above entry means that `acknowledge` is a verb that takes an IP, NP or CP as the complement. The lexicon is extensible in that users can define new functions to suit their own needs. Current implementation of the lexicon also includes functions `ref` and `phrase`, which are explained in the next two subsections.

### 4.2. Reference Entries

The lexicon does not contain separate entries for regular variations of words. When a word is not found in the lexicon, the lexical retriever strips the endings of the word to recover possible base forms of the word and look them up in the lexicon. For example, when the lexical retriever fails to find an entry for "studies," it searches the lexicon for "studie," "studi" and "study." Only the last one of these has an entry in the lexicon and its entry is returned.

Irregular variations of words are explicitly listed in the lexicon. For example, there is an entry for the word "began." However, the subcatgorization frames of "begin" are not listed again under "began." Instead, the entry contains a `ref` function which returns a reference to the entry for "begin."
```
(began
 (ref ((cat v) (vform ed) -prog -perf -passive
       (tense past))) (begin (cat))))
```
The first argument of `ref` is the attribute values of "began." The second argument contains the base form of the word and a set of attribute names. The lexical items for the word "began" is obtained by unifying its attribute values with the attribute values in the lexical entry for "begin." The advantage of making references to the base form is that when the base form is modified, one does not have to make changes to the entries for its variations.

### 4.3. Phrasal Entries

The lexicon also allows for phrases that consist of multiple words. One of the words in a phrase is designated as the head word. The head word should be a word in the phrase that can undergo morphological changes and is the most infrequent. For example, in the phrase, "down payment," the head word is "payment." In the lexicon, a phrase "$w_1 \ldots w_h \ldots w_n$" is stored as a string "$w_h \ldots w_n, w_1 \ldots w_{h-1}$." That is, the first word in the string is always head word and the words after "," should appear before the head word in texts. The function `phrases` converts its arguments into a list of phrases where the entry word is the head. For example, the lexical entry for "payment" is as follows:

```
(payment
 (subcat ((cat n) (nform norm)))
 (phrases
   (payment, down)
   (payment, stop)
   (payment, token)
   (payment, transfer)))
```

After retrieving the entry for a word, each phrase in the phrase list is compared with the surrounding words in the sentence. If the phrase is found in the sentence, the entry for the phrase is retrieved from the lexicon.

### 5. Reducing Ambiguities

One of the problems with many parsers is that they typically generate far more parses than humans normally do. For example, the average number of parses per word is 1.35 in (Black et al., 1992). That means that their parser produces, on average, 8 parses for a 7-word sentence, 34 parses for a 12-word sentence, and 144 parses for a 17-word sentence. The large number of parse trees make the processing at later stages more difficult and error prone.

PRINCIPAR defines a weight for every parse tree. A weight is associated with every word sense and every link in the parse tree. The weight of the parse tree is the total weight of the links and the word senses at the leaf nodes of the tree.

The packed shared parse forest in PRINCIPAR is organized in such a way that the parse tree with minimum weight is retrieved first. PRINCIPAR then uses the minimum weight and a predetermined number called BIGWEIGHT, which is currently arbitrarily defined to be 20, to prone the parse forest. Only the parse trees whose weights are less than (minimum weight + BIGWEIGHT/2) are spared and output.

The weights of the links and word senses are determined as follows:

- The links from Xbar to an adjunct YP have weight=BIGWEIGHT and all the other links have weight=1.0.

- The words in the lexicon may have an attribute `rare`, which takes values from {very, very-very}. If a word sense has the attribute value (rare very), its weight is BIGWEIGHT. If a word sense has the attribute value (rare very-very), its weight is 2×BIGWEIGHT. Otherwise, the weight is 0.

Note that the attribute `rare` is used to indicate the relative frequency among different senses of the same word.

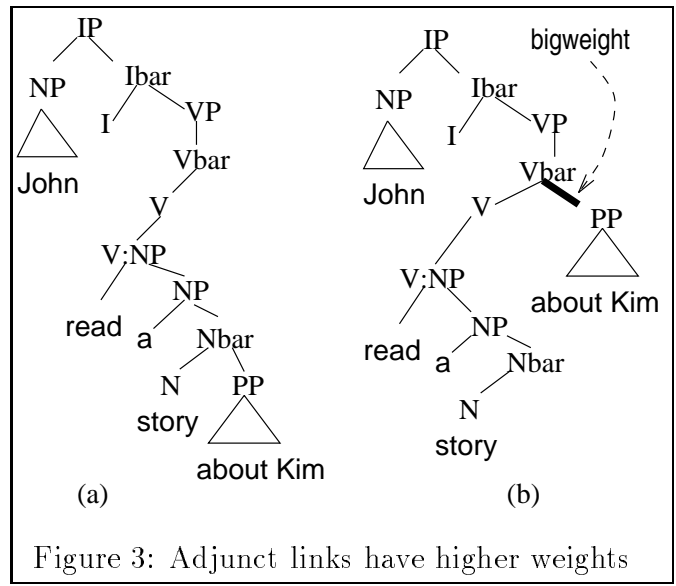

Figure 3: Adjunct links have higher weights

**Example 5.1.** Comparing the two parses of the sentence "John read the story about Kim" in Figure 3: in (a), [PP about Kim] is the complement of "story"; in (b), it is the adjunct of "read". Since the adjunct dominance link from Vbar to PP has much higher weight than the complement dominance link from Nbar to PP, the total weight of (a) is much smaller than the weight of (b). Therefore, only (a) is output as the parse tree of the sentence.

**Example 5.2.** The lexical entry for the word "do" is as follows:

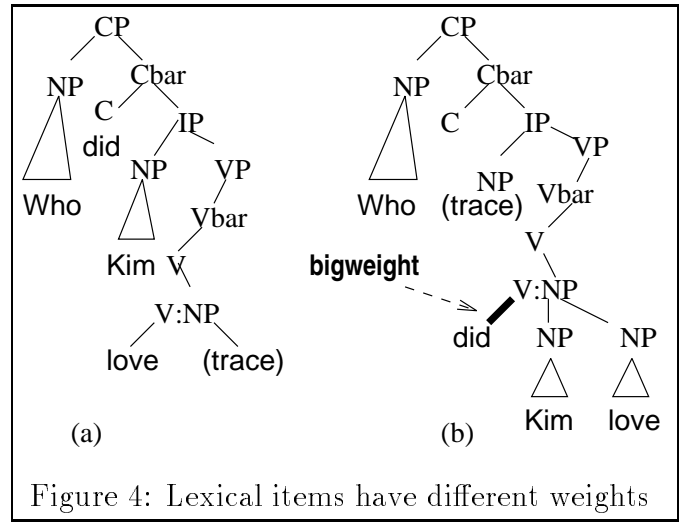

Figure 4: Lexical items have different weights

```
(do
 (subcat ((cat i) -passive -perf (auxform do)
          -prog (cform fin) (tense present)))
 (subcat ((cat v) (rare very))
         (((cat n) (case acc) (nform norm))))
 (subcat ((cat v) (rare very-very))
         (((cat n) (case acc) (nform norm))
          ((cat n) (case acc) (nform norm)))))
```

That is "do" can be an auxiliary verb, a transitive verb or a di-transitive verb. Figure 4 shows two parse trees for the sentence "Who did Kim love?" The parse tree (a) corresponds to the correct understanding of the sentence. In (b), "did" is analyzed as a bi-transitive verb as in "Who did Kim a favor?" However, since the latter sense of the word has an attribute value (`rare very-very`), tree (b) has much higher weight than tree (a) and only (a) is output by the parser.

## 6. Implementation and Experimental Results

PRINCIPAR has been implemented in C++. The graphical user interface is developed with a toolkit called InterViews. The program runs on SUN Sparcstations with X-windows. A version without graphical user interface can also be run on most Unix machines with GNU g++ compiler.

Table 1: Experimental Results

| Example sentences | words | time* | parses |
|---|---|---|---|
| Who do you think Bill said Mary expected to see | 10 | 0.46 | 1 |
| I asked which books he told me that I should read | 11 | 0.76 | 1 |
| The petition listed the mayor's occupation as attorney and his age as 71 | 13 | 0.60 | 14 |
| He said evidence was obtained in violation of the legal rights of citizens | 13 | 0.55 | 4 |
| Mr. Nixon , for his part , would oppose intervention in Cuba without specific provocation | 13 | 0.51 | 6 |
| The assembly language provides a means for writing a program and you are not concerned with actual memory addresses | 19 | 0.80 | 2 |
| Labels can be assigned to a particular instruction step in a source program that identify that step as an entry point for use in subsequent instructions | 26 | 4.13 | 32 |

* time (in seconds) taken on a Sparcstation ELC.

Lin and Goebel (1993) showed that the complexity of the message passing algorithm is $O(|G|n^3)$ for context-free grammars, where $n$ is the length of input sentence, $|G|$ is size of the grammar (measure by the number of the total length of the phrase structure rules). When attribute values are used in messages, the complexity of the algorithm is not yet known. Our experiments have shown that the parser is very fast. Table 1 lists the parsing time and the number of parses for several example sentences. The correct parses for all the sentences in Table 1 are returned by the parser. Even though the lexicon is derived from machine readable dictionaries and contains a large number of senses for many words, the ratio between the number of parse trees and the sentence length here is well bellow the ratio reported in (Black et al., 1992).

## Acknowledgements

The author wishes to thank Bonnie Dorr for comments about Sections 1, 2, and 3. This research was supported by Natural Sciences and Engineering Research Council of Canada grant OGP121338.